\documentclass[runningheads]{llncs}

\usepackage{comment}
\usepackage{graphicx}
\usepackage{caption}
\usepackage{booktabs}
\usepackage{multirow}
\usepackage{color}
\usepackage{pgfplots}
\usetikzlibrary{plotmarks}

\usepackage{booktabs} 
\usepackage{lineno}
\usepackage{amsmath}
\usepackage{amsfonts}
\usepackage{algorithm}
\usepackage{longtable}
\usepackage{multicol}
\usepackage{placeins}
 \usepackage{multirow}
\usepackage{hyperref}
\usepackage{eucal}

\usepackage{xspace}
\usepackage{graphicx}
\usepackage{paralist}
\usepackage{url}

\usepackage{amsfonts,amsmath,amssymb}

\usepackage[utf8]{inputenc}
\usepackage{graphicx}
\usepackage{verbatim}

\usepackage{multirow}
\usepackage{booktabs}
\usepackage{url}
\usepackage{subfig}
\usepackage{dblfloatfix} 
\usepackage{cite}

\usepackage{graphicx}
\usepackage{setspace}
\usepackage{multicol}

\usepackage{url}

\urldef{\mailsa}\path|toutouh@mit.edu|

\pgfplotsset{compat=1.15}

\begin{document}

\title{Conditional Generative Adversarial Networks to Model Urban Outdoor Air Pollution}
\author{Jamal Toutouh\inst{1}}
\titlerunning{}
\authorrunning{J. Toutouh} 
\tocauthor{J. Toutouh}
\institute{Massachusetts Institute of Technology, CSAIL, MA, 
USA \\
\mailsa
}

\maketitle

\begin{abstract}
This is a relevant problem because the design of most cities prioritizes the use of motorized vehicles,  which has degraded air quality in recent years, having a negative effect on urban health. Modeling, predicting, and forecasting ambient air pollution is an important way to deal with this issue because it would be helpful for decision-makers and urban city planners to understand the phenomena and to take solutions. In general, data-driven methods for modeling, predicting, and forecasting outdoor pollution requires an important amount of data, which may limit their accuracy. In order to deal with such a lack of data, we propose to train models able to generate synthetic nitrogen dioxide daily time series according to a given classification that will allow an unlimited generation of realistic data. The main experimental results indicate that the proposed approach is able to generate accurate and diverse pollution daily time series, while requiring reduced computational time. 
\end{abstract}

\section{Introduction}
\label{sec:introduction}

Artificial intelligence, computational intelligence, and automated learning are already coping with different areas in our daily life due to their success in a wide range of applications~\cite{Engelbrecht2007}. 
In this study, we focus on generative models, 
which have shown success on tasks related to learning and gaining knowledge about data, data distributions, and other valuable information~\cite{wang2019generative}.

In particular, generative adversarial networks (GANs) is a powerful method to train generative models~\cite{goodfellow2014generative}. 
GANs take a training set drawn from a specific distribution and learn to represent an estimate of that distribution by using unsupervised learning. 
The output of this method is a generative model that produces new information units that approximate the original training set.

In general, GANs consist of two artificial neural networks (ANN), a generator and a discriminator, that apply adversarial learning to optimize their parameters (weights). The discriminator learns how to distinguish between the ``natural/real'' data samples coming from the training dataset and the ``artificial/fake'' data samples produced by the generator. 
The generator is trained to deceive the discriminator by transforming its inputs from a random latent space into ``artificial/fake'' data samples. 
GAN training is formulated as a minimax optimization problem by the definitions of generator and discriminator loss~\cite{goodfellow2014generative}.

GANs have been successfully applied to generate realistic, complex, and multivariate distributions. This has motivated a growing body of applications, especially those concerning multimedia information (e.g., images, sound, and video), in science, design, art, games, and other areas~\cite{Pan2019,carla_2020}. 

Urban design has traditionally prioritized motorized mobility (the use of the individual or collective vehicles), with the growth of the cities this is having an undesired negative effect over safety and reducing the quality of life of the inhabitants. 
A major concern derived from the rapid development of car-oriented cities is the high generation of air pollutants and their impacts on the citizens' health~\cite{soni2016}.  
Thus, air pollution is the top health hazard in the European Union because it reduces life expectancy and diminishes the quality of health~\cite{Steele2001,lebrusan2020car}.

In the urban areas, one of the major sources of pollutants, such as nitrogen dioxide (NO$_2$), is road traffic~\cite{Steele2001}. Thus,  reducing it would be an effective strategy to improve urban livability and their inhabitants' health. 
However, it is not easy to understand the different phenomena that may have implications for the production or dissipation of pollutants. 
For this reason, there have been different approaches to evaluate the real impact of mobility policies in the air quality~\cite{lebrusan2019assessing, lebrusan2020sc, toutouh2020computational,sobkova2017urban}.

The interest in modeling, predicting, and forecasting ambient air pollution has also been growing during the last years. Getting in advance accurate quality air values would allow policy-makers and urban city planners to provide rapid solutions to prevent human risk situations~\cite{moustris20103}.  

Traditionally, physics-based and deterministic approaches have been applied to address air pollution modeling~\cite{mueller2011contributions,chuang2011application}.
These approaches 
are sensitive to several factors, including the scale and quality of the parameters involved, computationally expensive, and dependent on large databases of several input parameters, of which some may not be available~\cite{cabaneros2019review}.

With the rapid development of ANNs and their successful application to many different short-term and long-term forecasting applications, several researchers have proposed the use of such a data-driven methodology to deal with air outdoor pollution modeling, prediction, and forecasting~\cite{cabaneros2019review}. 
On the one hand, the main advantage of this approach is that the use of ANNs does not require an in-depth understanding of the physics and dynamics between air pollution concentration levels and other explanatory variables~\cite{qi2019hybrid,liu2019intelligent,cabaneros2019review}. 
On the other hand, it is an open question the selection of the appropriate ANN model, the interpretation of the results of that kind of black-box methods, and the results are problem specific. 
Besides, as the deterministic models, this kind of machine learning and deep learning methods require a vast amount of data to be trained.

In this research, we want to propose the use of a specific type of GANs, conditional GANs (CGANs)~\cite{mirza2014conditional}, to train generators able to create synthesized data, as a data augmentation approach, to feed data-driven methodologies for modeling, forecasting, and predicting outdoor pollution. 
The data samples generated are the daily time series of a given pollutant from a given area of a city according to a given condition (class). 
As a use case, we deal with the generation of daily NO$_2$ concentration time series at the \textit{Plaza the Espa\~{n}a} in Madrid (Spain). The real dataset provided used to train the CGAN is build by collecting the levels of NO$_2$ gathered by a sensor located there. It is important to remark that we are not trying to create a pollution forecasting method, but a modeling one from training the generative models. 

The main contributions of this research are: 
a) proposing a new approach based on CGANs to create pollution time series, 
and b) generating new data samples to be used by data-driven pollution modeling approaches. Thus, we want to answer the following research question: \textbf{RQ}: Is it possible to apply generative modeling to produce new daily time series to improve our understanding of the phenomena related to the pollution in our cities? 

The paper is organized as follows: 
The next section introduces the main concepts to understand CGANs and how they are applied to generate air pollution data.
Section~\ref{sec:methodology} introduces the research methodology applied in this research.
The experimental analysis is presented in Section~\ref{sec:results}. 
Finally, Section~\ref{sec:conclusions} draws the conclusions and the main lines of future work.

\section{CGANs for pollution data augmentation}
\label{sec:CGAN-pollution}
The CGANs are an extension of a GAN for conditional settings (labeled data).
This section introduces the main concepts in GANs and CGANs training and presents 
how CGANs are applied to generate pollution daily series to address pollution data augmentation. 

\subsection{Conditional generative adversarial networks training}

GANs are unsupervised learning methods that learn the specific distribution of a given (real) training dataset, to produce samples using the estimated distribution. 
Generally, GANs consist of a generator and a discriminator that apply adversarial learning to optimize their parameters. 

\begin{figure}[!h]
\centering
\begin{minipage}[l]{0.7\textwidth}
\centering
    \includegraphics[width=\textwidth, trim=2cm 2cm 5cm 2cm, clip]{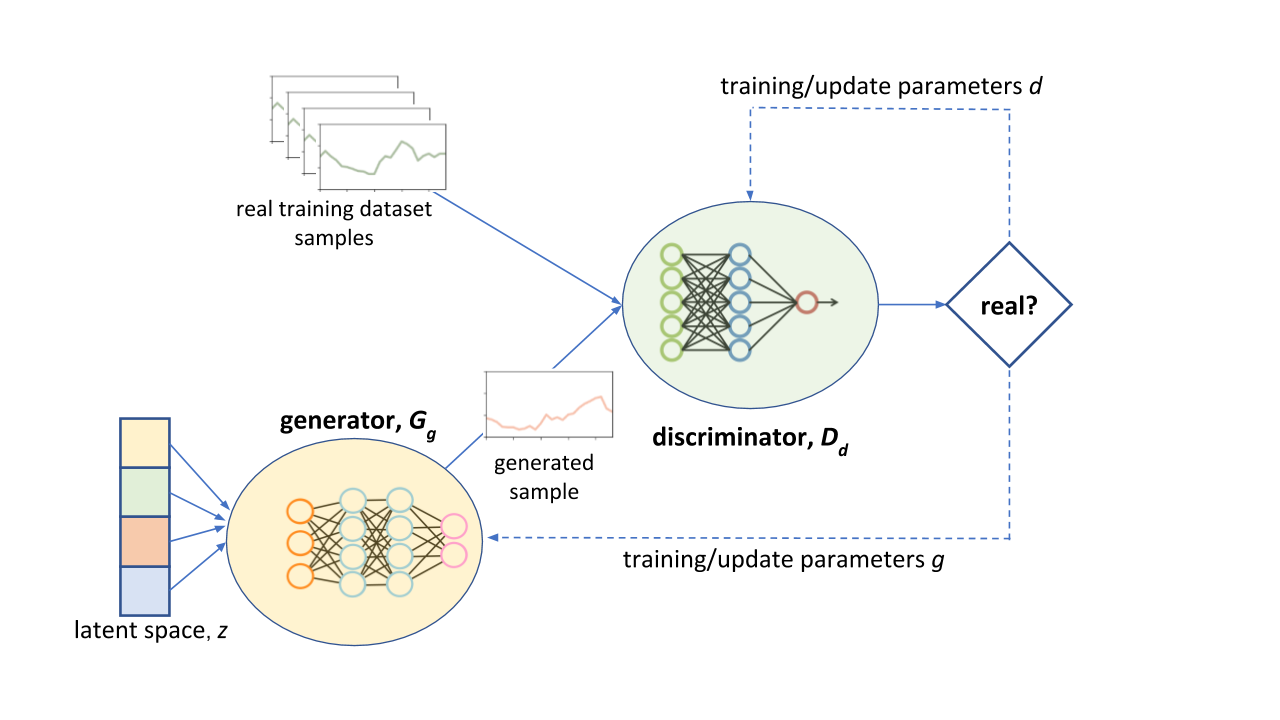}
    \footnotesize{a) GAN training.}
\end{minipage}
\vspace{0.2cm}

\begin{minipage}[l]{0.7\textwidth}
\centering
    \includegraphics[width=\textwidth, trim=2cm 2cm 5cm 2cm, clip]{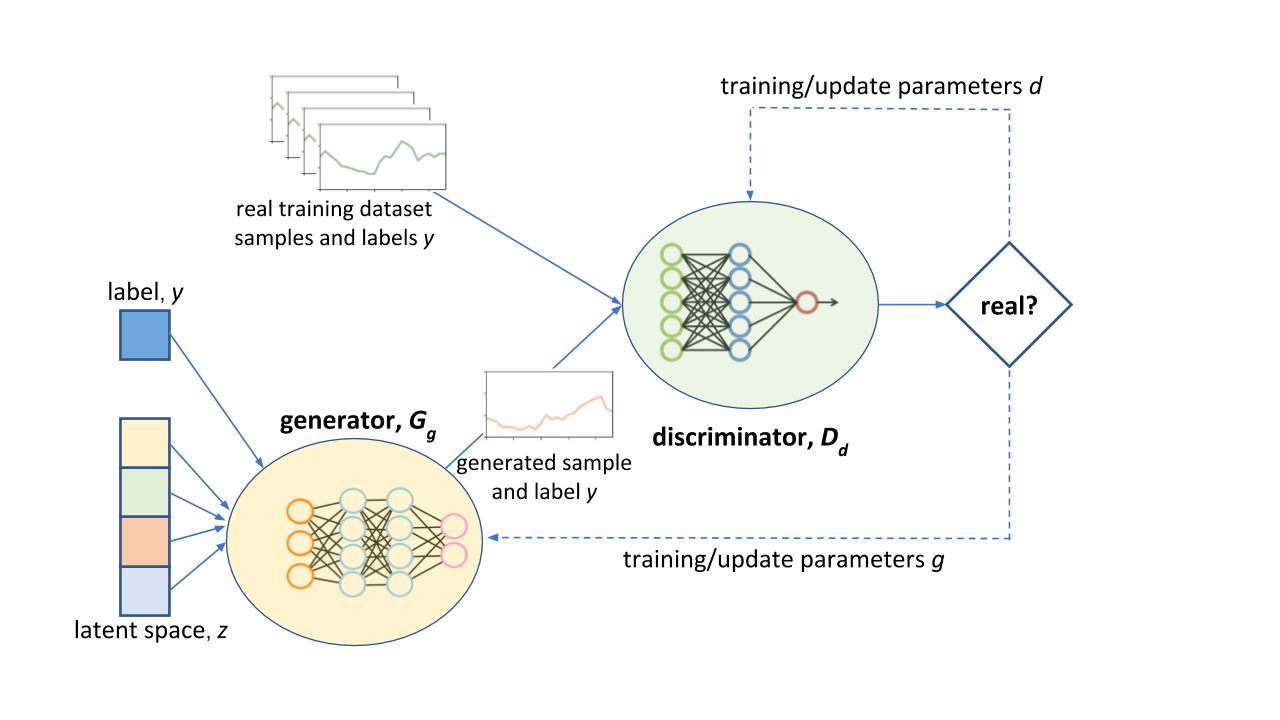}
\end{minipage}
\vspace{0.2cm}

\caption{Most commonly used CGANs training setups.}
\label{fig:gan-cgan}
\end{figure}

During the training process, the discriminator updates its parameters to learn how to differentiate between the natural/real samples from the training data set and the artificial/fake samples synthesized by the generator (see Fig.~\ref{fig:gan-cgan}).

The GAN training problem is formulated as a minimax optimization problem by the definitions of generator and discriminator. 
Let $\mathcal{G}=\{G_g,\ g \in \mathbb{G}\}$ and $\mathcal{D}=\{D_d,\ d \in \mathbb{D}\}$ denote the class of
generators and discriminators, where $G_g$ and $D_d$ are functions
parameterized by $g$ and $d$.  $\mathbb{G}, \mathbb{D} \subseteq \mathbb{R}^{p}$
represent the respective parameters space of the generators and
discriminators. 
The generators $G_g$ map a noise variable from a latent space $z\sim P_{z}(z)$ to data space $x = G_{g}(z)$. 
The discriminators $D_d$ assign a probability $p = D_{d}(x) \in [0, 1]$ to represent the likelihood that $x$ belongs to the real training data set. 
In order to do so, $\phi:[0,1] \to \mathbb{R}$, which is concave \emph{measuring function}, is used. 
The $P_{z}(z)$ is a prior on $z$ 
(a uniform $[-1, 1]$ distribution is typically chosen). 
The goal of GAN training is to find $d$ and $g$ parameters 
to optimize the objective function $\mathcal{L}(g,d)$.
\vspace{-0.2cm}
\begin{equation}
\label{eq:gan-def}
\min_{g\in \mathbb{G}}\max_{d \in \mathbb{D}} \mathcal{L}(g,d),\ \text{where} \nonumber
\vspace{-0.2cm}
\end{equation}
\begin{equation}
\mathcal{L}(g,d) = \mathbb{E}_{x\sim P_{data}(x)}[\phi (D_d(x))] + \mathbb{E}_{z\sim P_{z}(z)}[\phi(1-D_d(G_g(z)))]\;,
\end{equation}

This provokes that $D_d$ becomes into a binary classifier providing the best possible discrimination between real and fake
data. Simultaneously, it encourages $G_g$ to fit the real data
distribution. In general, both ANN are trained by applying backpropagation.


CGANs are an extension of GANs to deal with labeled training datasets (structured in classes). The idea is to train generative models able to create samples of a given class given according to a given label. Thus $G_g$ and $D_d$
receive an additional variable $y$ as input, which represents the label of the class (see Fig~\ref{fig:gan-cgan}.b). The CGANs objective function can be rewritten it is shown in Eq~\ref{eq:cgan}.

\vspace{-0.2cm}
\begin{equation}
\label{eq:cgan}
\min_{g\in \mathbb{G}}\max_{d \in \mathbb{D}} \mathcal{L}(g,d),\ \text{where} \nonumber
\vspace{-0.2cm}
\end{equation}
\begin{equation}
\mathcal{L}(g,d) = \mathbb{E}_{x,y\sim P_{data}(x, y)}[\phi (D_d(x, y))] + \mathbb{E}_{z\sim P_{z}(z), y\sim P_{y}(y)}[\phi(1-D_d(G_g(z,y), y))]\;,
\end{equation}

\vspace{-0.5cm}
\subsection{GANs applied to urban sciences}
\vspace{-0.2cm}

There is a consistent body of evidence that GANs excel at implicitly sampling from highly complex, analytically-unknown distributions in a great number of contexts~\cite{wang2019generative}.
Thus, nowadays, researchers are using such a machine learning approach to generate data across a variety of disciplines. 

However, there is a lack of studies on applying GANs to problems related to our cities. 
Some examples of this type of research are: 
a generative model trained by an unconditional GAN was trained to create realistic built land use maps, i.e., the model was able to generate cities (maps of built land use). These maps showed a high degree of realism and they provided realistic values on several statistics used in the urban modeling literature with real cities~\cite{albert2018modeling}. 
Later, the same authors proposed the use of CGANs to add new levels of realism to the generated maps and the ability to predict land use maps from underlying socio-economic factors.
These new maps take into account, for example, physical constraints such as water areas~\cite{albert2019spatial}.
Physics-informed GAN (PIGAN) was proposed to enhance the performance of GANs by incorporating both constraints of covariance structure and physical laws. 
The idea es to improve the robustness of GANs when dealing with problems related to remote sensing~\cite{paper_59}.
This is important for applications that require the use of satellite data and could suffer from phenomena like the appearance of clouds.
In turn, we could include in this class of urban sciences research by applying GANs, 
the study of applying GANs to create synthetic data about building energy consumption in order to be used together with real data to train a data-driven forecasting model to predict energy consumption~\cite{TIAN2019230}. The idea was to overcome the issues of having a lack of data to train the model to predict with enough accuracy. 

\vspace{-0.3cm}
\subsection{Pollution daily time series generation}
\label{sec:pollution-generation}
\vspace{-0.1cm}

\begin{figure}
\vspace{-0.6cm}
\centering
    \includegraphics[trim=0 5.4cm 0 0, clip, width=0.85\textwidth]{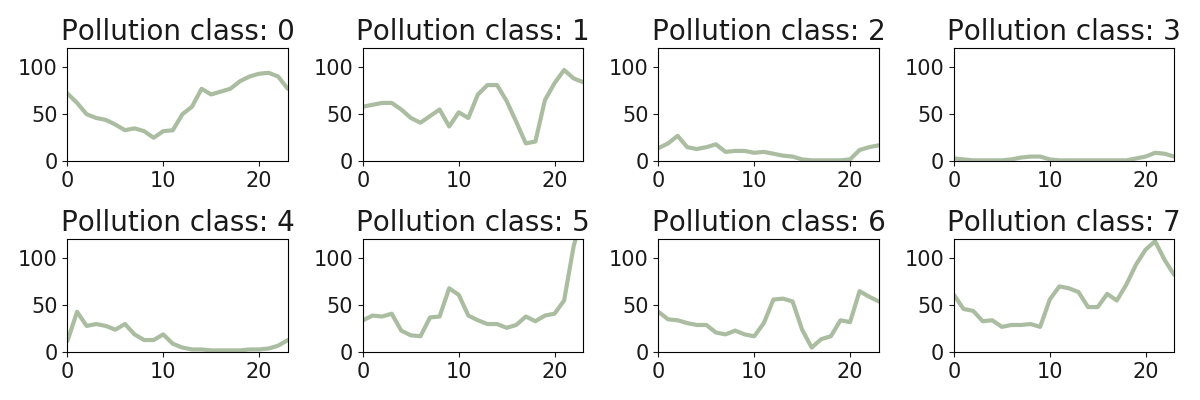}
    \vspace{-0.2cm}
\caption{Some samples of the real dataset, i.e., NO$_2$ pollution daily series.}
\label{fig:real-data-samples}
\vspace{-1.cm}
\end{figure}

Dealing with the problem of not having enough data to analyze pollution in our cities, we proposed an approach for data augmentation to create synthetic daily time series of given pollutant. 
The approach consists in sampling from an existing dataset of real daily time series of given pollutant labeled according to different classes to train the discriminator, 
while the generator reads a vector $z$ (from the latent space) and a label $y$ to generate a fake daily time series (see Fig.~\ref{fig:gan-cgan}). 
The generator and discriminator are trained against each other. 
Fig.~\ref{fig:real-data-samples} illustrates samples of the real training dataset.

\vspace{-0.3cm}

\section{Materials and methods}
\label{sec:methodology}
\vspace{-0.2cm}

This section presents the applied methodology for training generative models by using GANs to create synthetic pollution data.

\vspace{-0.3cm}
\subsection{Training dataset description}
\vspace{-0.1cm}

The training dataset studied here is provided by the Open Data Portal (ODP) offered by the Madrid City Council\footnote{Madird Open data Portal web - \url{https://datos.madrid.es/}}.
Specifically, the training dataset is the NO$_2$ concentration gathered by a sensor located at \textit{Plaza de Espa\~na}, which is in the downtown of the city. 
The dataset is built considering a temporal frame of five years, from January 2015 to December 2019. 
A given data sample is a time series that represents the NO$_2$ concentration of a given day which is averaged every hour.  
(Fig.~\ref{fig:real-data-samples}). Therefore, it could be seen as a vector of 24 continuous values.

Following previous research about the pollution in Madrid, the daily NO$_2$ concentration is classified into eight classes according to the season (winter, spring, summer, and autumn) and the type of day (i.e., working days, from Mondays to Fridays, and weekends, Saturdays and Sundays)~\cite{lebrusan2020car,lebrusan2020sc}. 
This classification follows the idea that warmer seasons have lower NO$_2$ concentration due to meteorological reasons and weekends have better air quality because the road traffic is lower than in working days. 
Table~\ref{tab:classes} presents this pollution classification and the number of samples per class. 


\begin{table}[!h]
\vspace{-0.7cm}
\setlength{\tabcolsep}{8pt}
\renewcommand{\arraystretch}{0.9}
\centering
\small
\caption{NO$_2$ pollution classification and number of samples per class.}
\label{tab:classes}
\begin{tabular}{llrr}
\toprule
season & type of day & class & number of samples\\
\midrule
winter & weekend & 0 & 439 \\ 
winter & working day & 1 & 1082 \\
spring & weekend & 2 & 439\\ 
spring & working day & 3 & 1119\\
summer & weekend & 4 & 445\\ 
summer & working day & 5 & 1116\\
autumn & weekend & 6 & 420 \\ 
autumn & working day & 7 & 1045 \\
\bottomrule
\end{tabular}
\vspace{-0.4cm}
\end{table}

The classes are highly unbalanced (see Table~\ref{tab:classes}), i.e., autumn-weekend (class 6) has 420 samples (the minimum) and spring-working day (3) has 1119  (the maximum). For our experiments, we randomly sampled over the classes to select 420 samples of each class to balance the dataset to avoid training biases. Thus, the training dataset size is 3360 (420$\times$8).   

\begin{figure}[!h]
\centering
\begin{minipage}[l]{0.45\textwidth}
\centering
    \includegraphics[width=\textwidth]{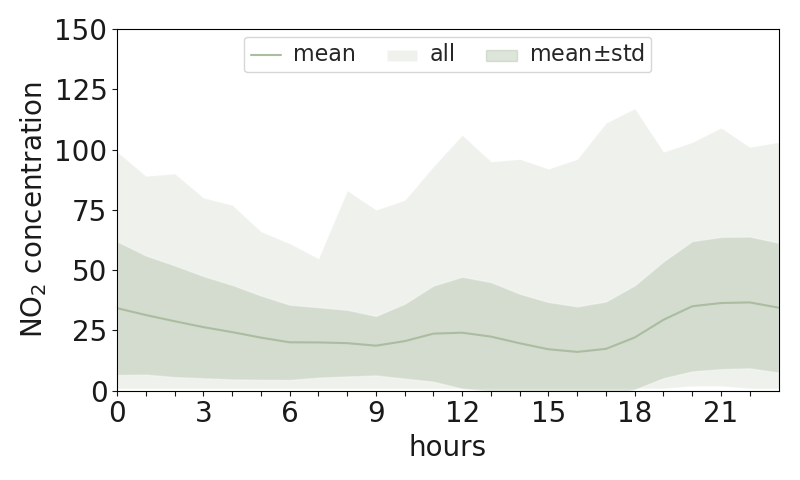}
    \vspace{-0.3cm}
    
    \footnotesize{a1) Class 0: winter-weekends.}
\end{minipage}
\begin{minipage}[l]{0.45\textwidth}
\centering
    \includegraphics[width=\textwidth]{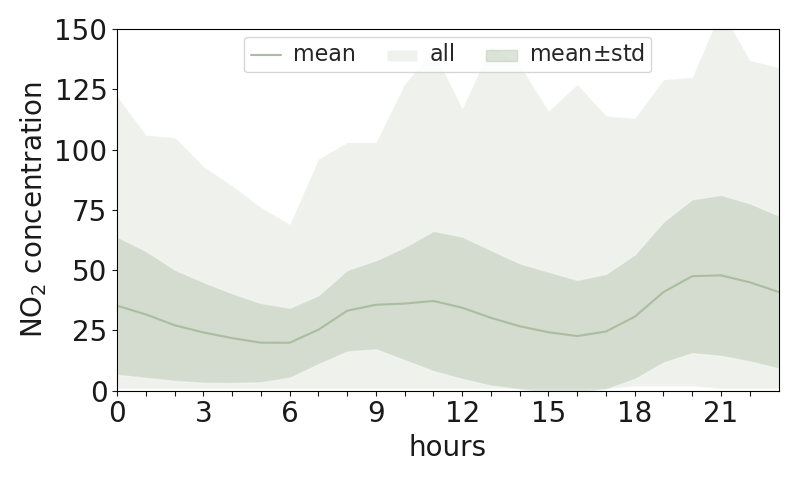}
        \vspace{-0.3cm}
        
    \footnotesize{a2)  Class 1: winter-working.} 
\end{minipage}

\begin{minipage}[l]{0.45\textwidth}
\centering
    \includegraphics[width=\textwidth]{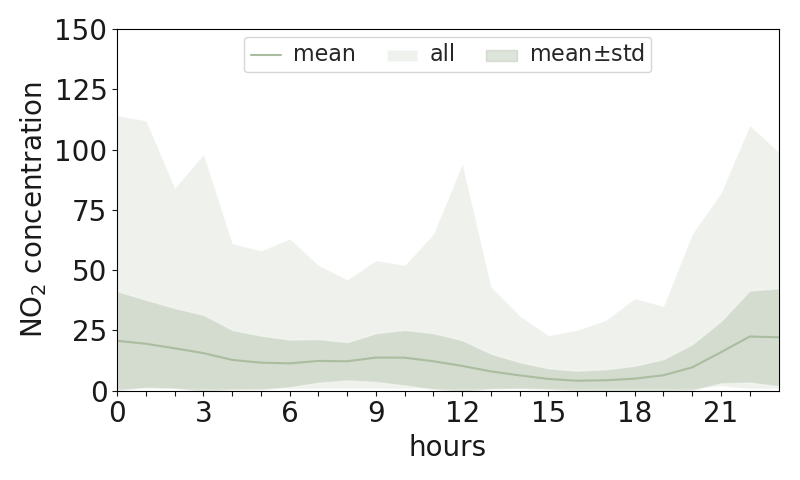}
        \vspace{-0.3cm}

\footnotesize{b1) Class 2: spring-weekends.}
\end{minipage} 
\begin{minipage}[l]{0.45\textwidth}
\centering
    \includegraphics[width=\textwidth]{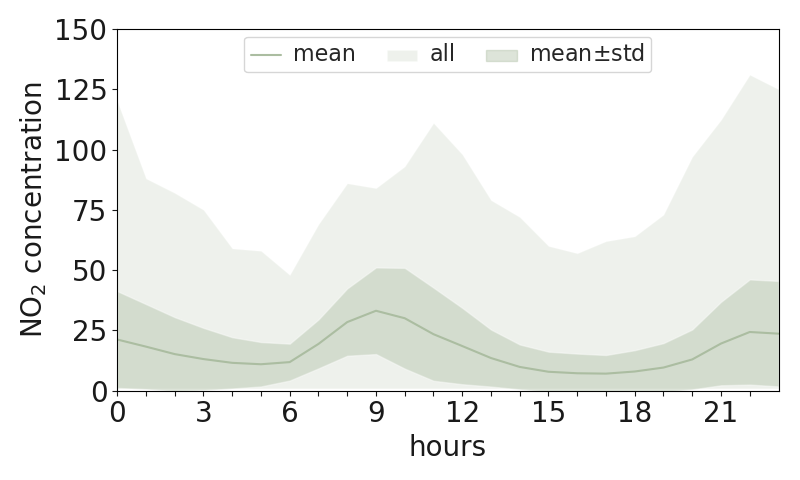}
       \vspace{-0.3cm}
       
 \footnotesize{b2) Class 3: spring-working.} 
\end{minipage} \\

\begin{minipage}[l]{0.45\textwidth}
\centering
    \includegraphics[width=\textwidth]{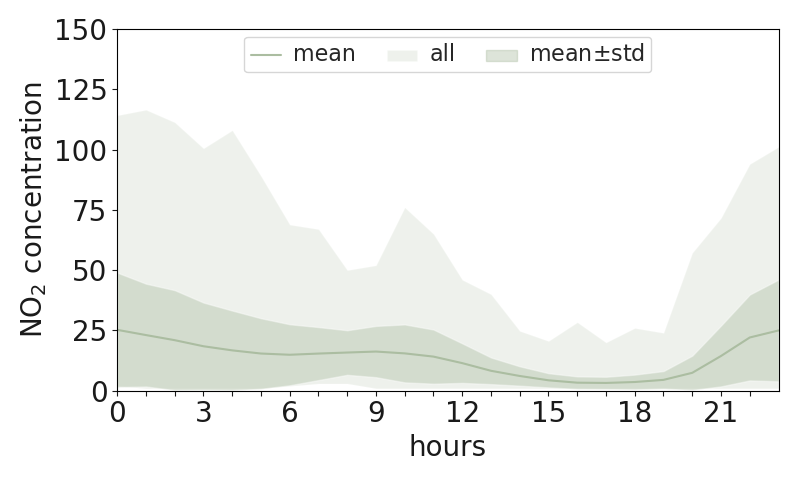}
        \vspace{-0.3cm}
        
\footnotesize{c1) Class 4: summer-weekends.}
\end{minipage} 
\begin{minipage}[l]{0.45\textwidth}
\centering
    \includegraphics[width=\textwidth]{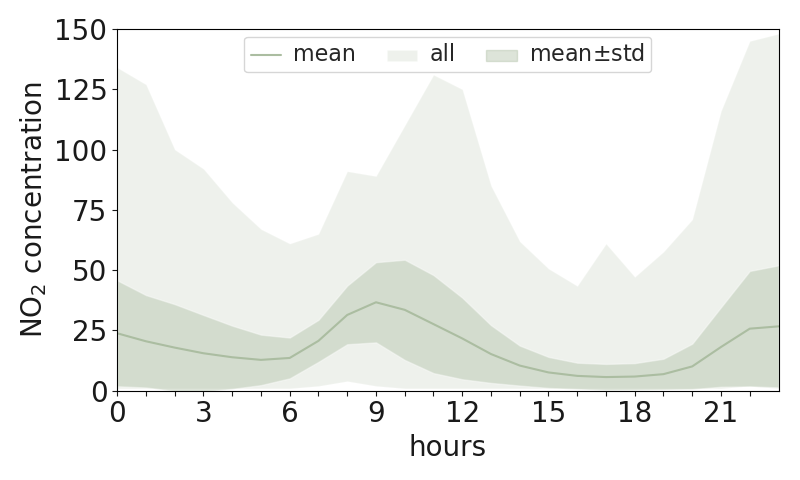}
    \vspace{-0.3cm}
    
    \footnotesize{c2) Class 5: summer-working.} 
\end{minipage} 

\begin{minipage}[l]{0.45\textwidth}
\centering
    \includegraphics[width=\textwidth]{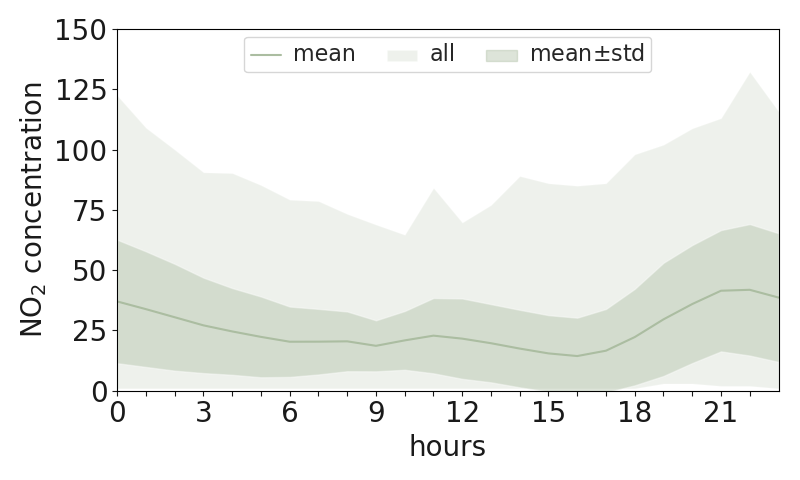}
        \vspace{-0.2cm}
        
\footnotesize{d1) Class 6: autumn-weekends.}
\end{minipage} 
\begin{minipage}[l]{0.45\textwidth}
\centering
    \includegraphics[width=\textwidth]{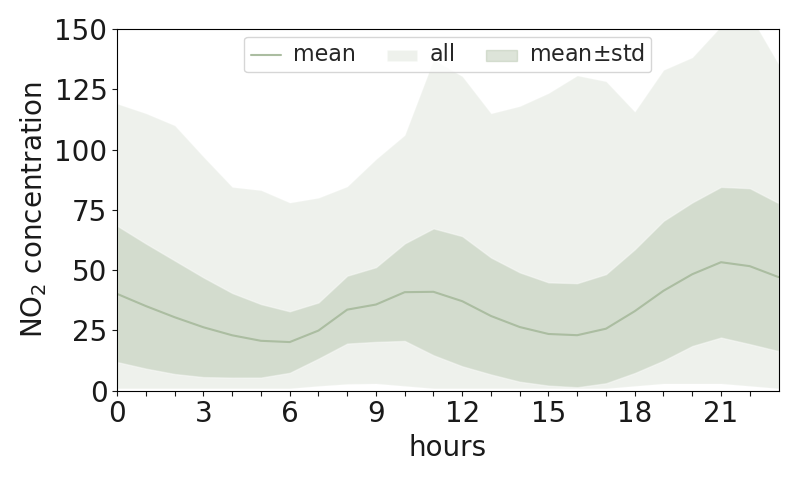}
        \vspace{-0.3cm}
        
\footnotesize{d2) Class 7: autumn-working.} 
\end{minipage} 

\caption{NO$_2$ hourly values for the seven classes defined in our study (real data).}
\label{fig:real_data}
\vspace{-0.5cm}
\end{figure}

Fig~\ref{fig:real_data} illustrates the training dataset. 
The green line shows the mean value, i.e., it contains the mean pollution values for the whole data of the class (we named it as the representative time series of a given class $c$, i.e., $rep_c$). 
The dark green area represents the values between the border defined by the mean minus the standard deviation and the mean plus the standard deviation (mean $\pm$ the standard deviation). 
Finally, the lighter green area represents the values between the minimum and maximum pollution measured in a given time.

\subsection{CGAN design details}

In our research, both ANNs, the generator and the discriminator, are implemented as multilayer perceptrons (MLP)~\cite{Hastie2009}. 
MLP are comprised of perceptrons or neurons, organized on layers. 
The minimum setup is formed by an input layer, which receives the problem data as input, and an output layer, which produces the results. 
In between, one or more hidden layers can be included to provide different levels of abstraction to help with the learning goal. 
The main difference between linear perceptrons and MLP is that all the neurons on the hidden and output layer apply a nonlinear activation function. 
MLP have shown competitive results when dealing with different kinds of machine learning, such as classification/prediction problems with labeled inputs, regression problems, etc. 

Our approach explores the use of a four-layer MLP to build the generator and the discriminator.  
The input of the generator has a size of 64 (size of $z$) plus eight (size of $y$) to specify the label of the class of the data sample to be generated, i.e., the total input size is 72.  
The output of the generator has the same input of the discriminator, which in this case is 24 (the size of the generated sample) plus eight (size of $y$) to identify the label of the sample, i.e., the total size is 32. 
Both types of MLP use linear layers. Hidden layers apply the leaky version of a rectified linear unit (LeakyRelu) as activation function, and the discriminator output layer the sigmoid function. 
The output of the generator applies a linear function since the output is in the range of real values, $\mathbb{R}$. The two hidden layers have 256 unites for both trained ANN models.

\subsection{Metrics evaluated}

As we are dealing with GAN training, we evaluate the loss values computed for the generator and the discriminator during the training process. 
The function applied to compute the loss is the binary cross-entropy (BCE)~\cite{goodfellow2014generative} ($\mathcal{L}_d$ and $\mathcal{L}_g$).
  

\begin{equation}
\mathcal{L}_d = \frac{1}{2} \mathbb{E}_{x,y\sim P_{data}(x, y)}[log (D_d(x, y))] - \frac{1}{2} \mathbb{E}_{z\sim P_{z}(z), y\sim P_{y}(y)}[log(1-D_d(G_g(z,y), y))]\;,
\end{equation}
\begin{equation}
\label{ec:g-loss}
\mathcal{L}_g = \mathbb{E}_{z\sim P_{z}(z), y\sim P_{y}(y)} [log(1-D_d(G_g(z,y), y))]
\end{equation}

When working with GANs it is important to assess the quality of the generated samples. 
In our problem, the aim is to generate accurate daily time series of pollution that follow the same distribution as the real pollution in \textit{Plaza de Espa\~na}. 
We propose the use of the root mean squared error (RMSE) between the fake samples produced and the time series that represents the mean ($rep_c$). 
Thus, given a fake sample of a given class $c$, i.e., $f_c$, the quality of the sample is given by the RMSE($rep_c$,$f_c$) shown in Eq.~\ref{eq:rmse}, where $rep_c(t)$ and $f_c(t)$ represent the pollution level at a given time $t$ of the representative time series $c$ and the fake sample $f_c$, respectivelly. Thus, lower values indicate better sample quality.

\begin{equation}
\label{eq:rmse}
RMSE(rep_c, f_c) = \sqrt{\frac{1}{24} \sum_{t \in [0, 23]} (rep_c(t) - f_c(t) )^2}  
\end{equation}

Finally, we also take into account the computational time in order to evaluate the cost of the proposed genetative method. 

\section{Experimental analysis}
\label{sec:results}
\vspace{-.2cm}

This section presents the numerical analysis of the proposed approach.

\vspace{-.4cm}

\subsection{Development, training configuration, and execution platform}
\vspace{-.2cm}

The proposed generative approach is implemented 
in Python3 using Pytorch as the main library to deal with ANNs (\url{pytorch.org}). 
The GAN training has configured with a learning rate of 0.0002, batch size of 16 samples (210 batches per training epoch), and 2000 iterations (training epochs).

The experiments have been performed on a workstation equipped with an Intel Core i7-7850H processor, 32GB of RAM memory, 100 GB of SSD storage for temporary files, and a Nvidia Tesla P100 GPUs with 12\~GB of memory. 

\vspace{-.3cm}

\subsection{Experimental results}
\vspace{-0.1cm}

This subsection reports the experimental results. The training process has been performed 10~times. 
Thus this section studies: \textit{a)} the accuracy of the generators is analyze, \textit{b)} the evolution of the loss during the training process, \textit{c)} the samples sythesized by the generators, and \textit{d)} the computational time.  

\vspace{-0.1cm}
\subsubsection{Accuracy of the trained generators}
\vspace{-0.1cm}

The accuracy of the generators is evaluated according to the RMSE between the created samples of a given class $c$ and the daily time series that represents the mean of that $c$ class. 
Thus, after training each generator, we create 40,000 samples (5,000 samples of each class) as fake datasets. 
Table~\ref{tab:rmse-total} shows the minimum, mean, standard deviation (stdev), and maximum of the calculated RMSE. 
The first row includes the same values obtained when computed the RMSE taking into account real samples. 
The fake datasets in the table are ranked according to their mean RMSE, thus \textit{fake-1} dataset contains the samples with the best quality (lowest RMSE) and fake-10 the samples with the highest RMSE.

\begin{table}[!h]
\setlength{\tabcolsep}{5pt}
\renewcommand{\arraystretch}{1.05}
\centering
\small
\caption{RMSE results for each dataset.}
\label{tab:rmse-total}
\begin{tabular}{lrrr}
\toprule
dataset & minimum & mean$\pm$stdev & maximum \\
\midrule
real & 3.6 & 17.3$\pm$8.5 & 67.0 \\ 
\midrule
fake-1 & 4.0 & 15.3$\pm$7.8 & 75.8 \\ 
fake-2 & 4.2 & 15.4$\pm$8.1 & 82.1 \\ 
fake-3 & 3.7 & 15.4$\pm$7.7 & 71.0 \\ 
fake-4 & 3.5 & 15.5$\pm$8.3 & 91.7 \\ 
fake-5 & 3.8 & 15.5$\pm$7.9 & 75.6 \\ 
fake-6 & 3.3 & 15.5$\pm$8.2 & 77.3 \\ 
fake-7 & 4.3 & 15.6$\pm$7.9 & 68.3 \\ 
fake-8 & 3.8 & 15.7$\pm$8.3 & 81.4 \\ 
fake-9 & 3.3 & 15.7$\pm$8.4 & 80.2 \\ 
fake-10 & 4.0 & 15.7$\pm$8.2 & 83.7 \\ 
\bottomrule
\end{tabular}

\end{table}

All the fake datasets present lower mean RMSE than the samples of the real dataset. This is mainly due that the training process converges to a generator that creates samples with limited diversity (real data shows the highest standard deviation). Even the results shown by our approach are competitive, it would be desirable to add diversity to the produced samples. This is still an open question that some authors are facing by providing generative models as a mixture of several generators~\cite{Toutouh_GECO2020}.

Table~\ref{tab:rmse-class} shows the mean and standard deviation (stdev) of the computed RMSE taking into account the classes. 
Real data samples show the highest differences for the eight classes. Besides, it can be seen that the classes that represent working days (classes represented by odd numbers) show the highest differences, for both real and fake datasets. This mainly indicates that there is not a general behavior that defines all the working days and therefore, we should take into account different kinds of classification to deal with working days (maybe taking into account the day itself).

\begin{table}[!h]
\setlength{\tabcolsep}{2pt}
\renewcommand{\arraystretch}{1.05}
\centering
\small
\caption{RMSE per class for each dataset.}
\label{tab:rmse-class}
\begin{tabular}{lrrrrrrrr}
\toprule
dataset & 0 & 1 & 2 & 3 & 4 & 5 & 6 & 7 \\
\midrule
real & 20.2$\pm$7.1 & 23.8$\pm$8.3 & 10.7$\pm$5.6 & 13.2$\pm$5.9 & 11.5$\pm$6.3 & 13.4$\pm$6.6 & 18.3$\pm$6.4 & 22.4$\pm$7.6 \\ 
\midrule
fake-1 & 17.6$\pm$9.1 & 20.1$\pm$9.0 & 10.5$\pm$5.4 & 11.2$\pm$5.1 & 10.1$\pm$4.1 & 11.1$\pm$3.9 & 15.3$\pm$6.5 & 19.3$\pm$7.2 \\
fake-2 & 17.6$\pm$9.8 & 19.8$\pm$8.8 & 10.1$\pm$4.7 & 10.7$\pm$4.3 & 9.5$\pm$3.3 & 11.1$\pm$4.2 & 15.9$\pm$7.6 & 19.5$\pm$7.7  \\ 
fake-3 & 17.7$\pm$8.7 & 20.7$\pm$9.0 & 9.7$\pm$4.5 & 10.7$\pm$4.0 & 9.7$\pm$3.4 & 10.8$\pm$3.4 & 15.7$\pm$6.1 & 19.9$\pm$6.9 \\ 
fake-4 & 18.6$\pm$9.3 & 20.6$\pm$9.5 & 9.9$\pm$5.1 & 10.6$\pm$3.9 & 9.4$\pm$3.8 & 10.7$\pm$3.6 & 15.6$\pm$7.3 & 20.1$\pm$7.9 \\ 
fake-5 & 17.9$\pm$8.6 & 20.2$\pm$9.1 & 9.8$\pm$4.2 & 11.0$\pm$4.4 & 10.0$\pm$3.7 & 11.2$\pm$4.1 & 15.3$\pm$6.6 & 19.9$\pm$7.3 \\ 
fake-6 & 18.6$\pm$9.9 & 20.5$\pm$8.7 & 9.3$\pm$5.1 & 10.9$\pm$5.0 & 9.1$\pm$3.0 & 11.0$\pm$3.8 & 15.4$\pm$6.6 & 20.1$\pm$7.8 \\ 
fake-7 & 18.1$\pm$9.7 & 20.3$\pm$8.2 & 10.4$\pm$4.7 & 11.4$\pm$5.6 & 10.0$\pm$3.7 & 11.8$\pm$4.5 & 15.5$\pm$6.4 & 19.4$\pm$7.3 \\ 
fake-8 & 18.4$\pm$10.0 & 20.9$\pm$9.5 & 10.2$\pm$5.2 & 11.5$\pm$5.6 & 10.0$\pm$3.9 & 11.4$\pm$4.2 & 15.9$\pm$6.6 & 19.6$\pm$8.0  \\ 
fake-9 & 18.5$\pm$10.7 & 20.0$\pm$9.0 & 9.7$\pm$5.1 & 11.3$\pm$4.8 & 9.6$\pm$3.6 & 11.1$\pm$4.4 & 15.6$\pm$6.8 & 19.3$\pm$7.7 \\ 
kake-10 & 18.0$\pm$8.9 & 20.8$\pm$9.6 & 10.7$\pm$5.1 & 11.3$\pm$4.9 & 9.8$\pm$4.1 & 11.4$\pm$4.3 & 15.7$\pm$6.5 & 19.7$\pm$8.0 \\ 
\bottomrule
\end{tabular}

\end{table}

\subsubsection{Training process}

\begin{figure}[!h]
\centering
\begin{minipage}[l]{0.48\textwidth}
\centering
    \includegraphics[width=\textwidth]{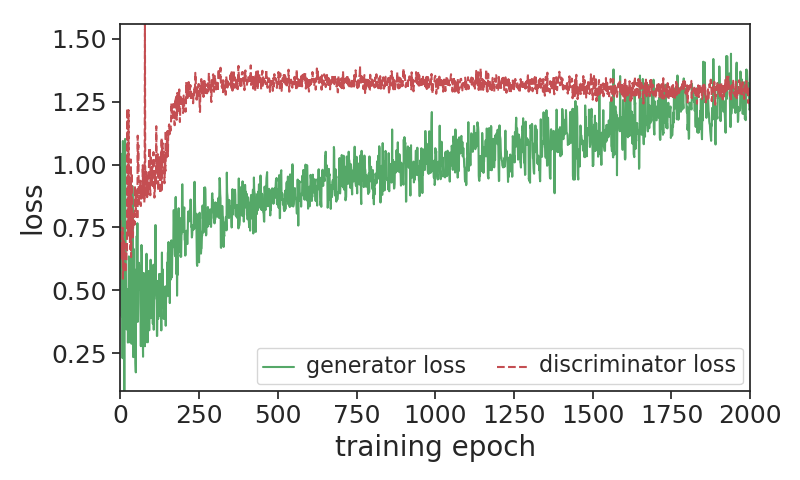}
    
    \vspace{-0.1cm}
    
    \footnotesize{a1) \textit{fake-1} loss values. }
\end{minipage}
\begin{minipage}[l]{0.48\textwidth}
\centering
    \includegraphics[width=\textwidth]{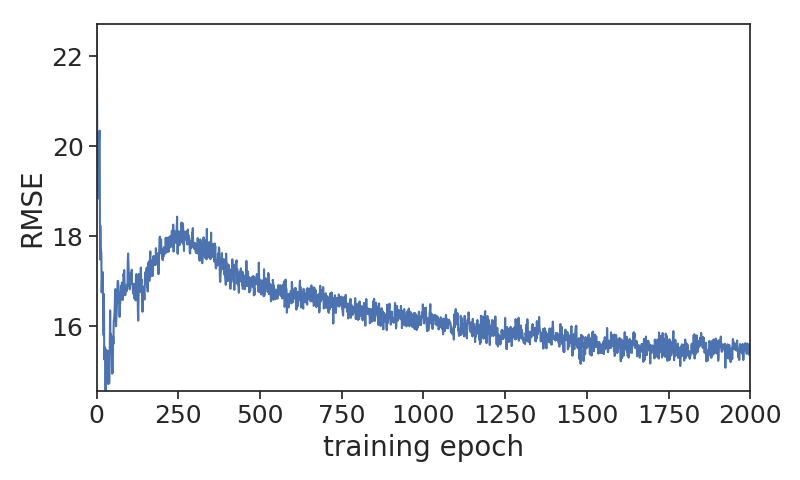}
    
    \vspace{-0.1cm}
    
    \footnotesize{a2) \textit{fake-1} RMSE.} 
\end{minipage}
\vspace{0.2cm}

\begin{minipage}[l]{0.48\textwidth}
\centering
    \includegraphics[width=\textwidth]{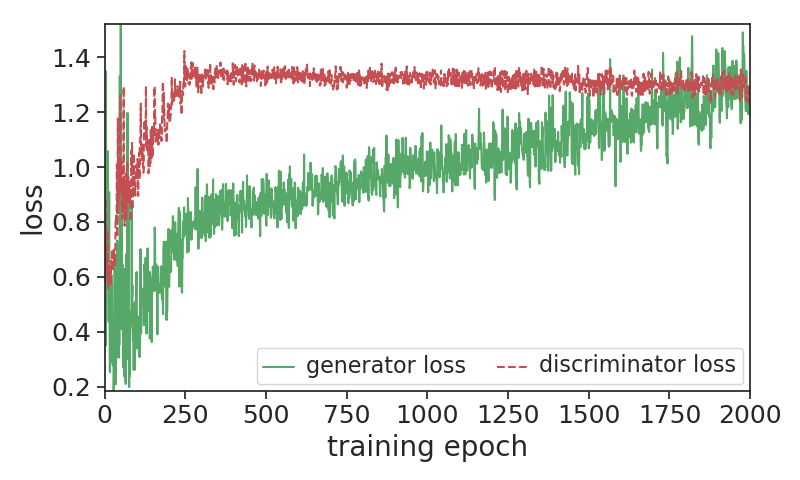}
    
    \vspace{-0.1cm}
    
    \footnotesize{b1) \textit{fake-5} loss values. }
\end{minipage} 
\begin{minipage}[l]{0.48\textwidth}
\centering
    \includegraphics[width=\textwidth]{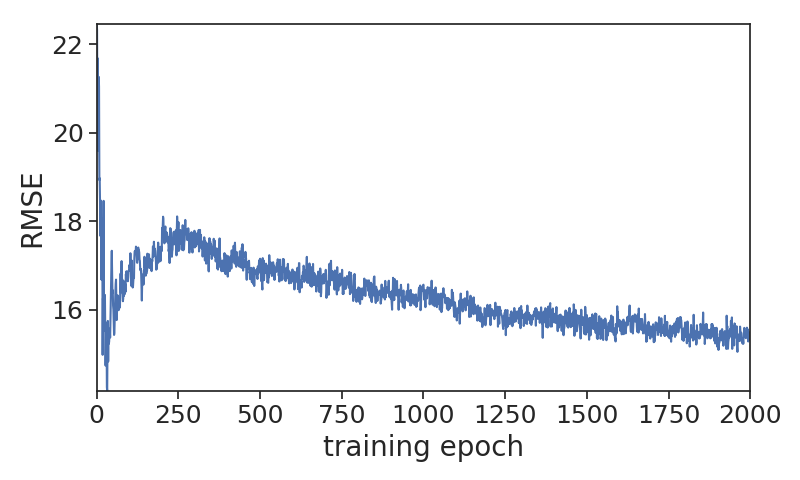}
    
    \vspace{-0.1cm}
    
    \footnotesize{b2) \textit{fake-5} RMSE.} 
\end{minipage} \\
\vspace{0.2cm}

\begin{minipage}[l]{0.48\textwidth}
\centering
    \includegraphics[width=\textwidth]{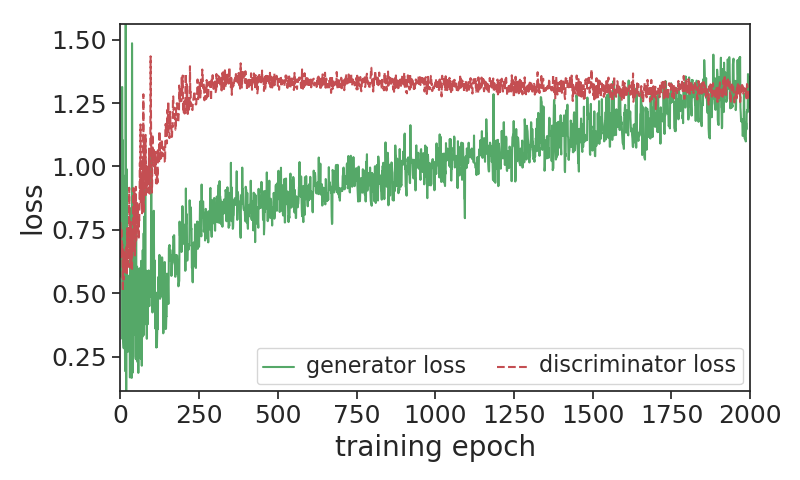}
    
    \vspace{-0.1cm}
    
    \footnotesize{c1) \textit{fake-10} loss values. }
\end{minipage} 
\begin{minipage}[l]{0.48\textwidth}
\centering
    \includegraphics[width=\textwidth]{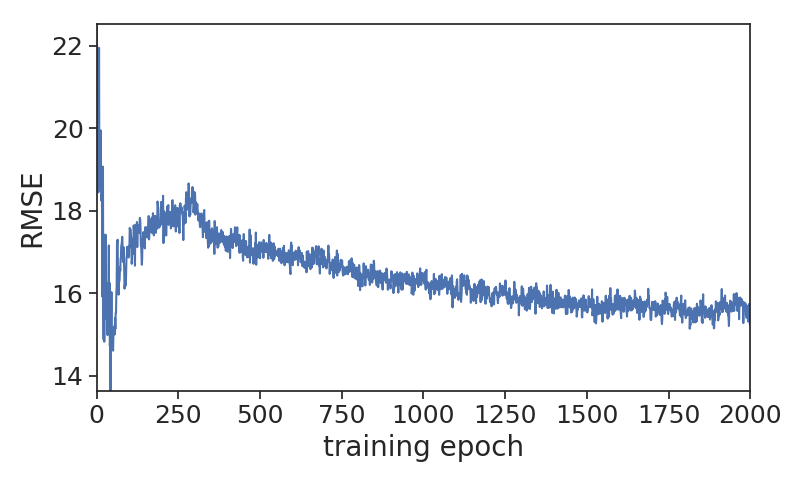}
    
    \vspace{-0.1cm}
    
    \footnotesize{c2) \textit{fake-10} RMSE.} 
\end{minipage} 
\caption{Discriminator and generator loss values and RMSE for three runs.}
\label{fig:training}
\end{figure}

In this section, we evaluate the behavior of the training process by showing the losses of the generator and the discriminator. In turn, we show the mean RMSE of the generated data at the end of each training epoch. 
The mean RMSE score is computed by creating 10,000 samples.
Fig.~\ref{fig:training} illustrates these metrics for the best, median, and worst run, i.e., \textit{fake-1}, \textit{fake-5}, and \textit{fake-10}, respectivelly.

Focusing on the evolution of the losses in Fig.~\ref{fig:training} a1, b1, and c1, we observe that the general behavior is very similar for the three evaluated runs. 
At the beginning (about the first 200-300~training epochs) the discriminator and the generator losses oscillate without showing a clear trend. This is mainly due to the discriminator has not been trained enough yet, and it is not able to discriminate anyhow between fake and real samples and it randomly assigns a loss value (i.e., it basically flips a coin). Thus, the generator can easily get low losses values that do not give the feedback required to learn (to update its parameters) property. For this reason, the samples produced provide increasing and oscillating RMSE (Fig.~\ref{fig:training} a2, b2, and c2). 

After that, the discriminator starts becoming stronger and reduces the loss values. Thus, it is harder for the generator to deceive the discriminator. For this reason, the generator starts increasing the computed loss values, allowing the generator to learn how to create more accurate samples. 
Therefore, the RMSE values start being reduced in an almost monotonically decreasing way.

After the 2,000 iterations, the generator and the discriminator seem to be in an equilibrium in which both loss values are similar. However, long runs will be able to give more insights into this concern.

It is important to remark that none of the runs shows typical GAN training pathologies, such as mode collapse, vanishing gradient, and oscillation; which would require to use more advanced GAN training methods, e.g., Lipizzaner~\cite{Schmiedlechner2018} or Mustangs~\cite{Toutouh2019}.

\subsubsection{Synthetic daily pollution time series generated}

\begin{figure}
\centering
\begin{minipage}[l]{0.48\textwidth}
\centering
    \includegraphics[width=\textwidth]{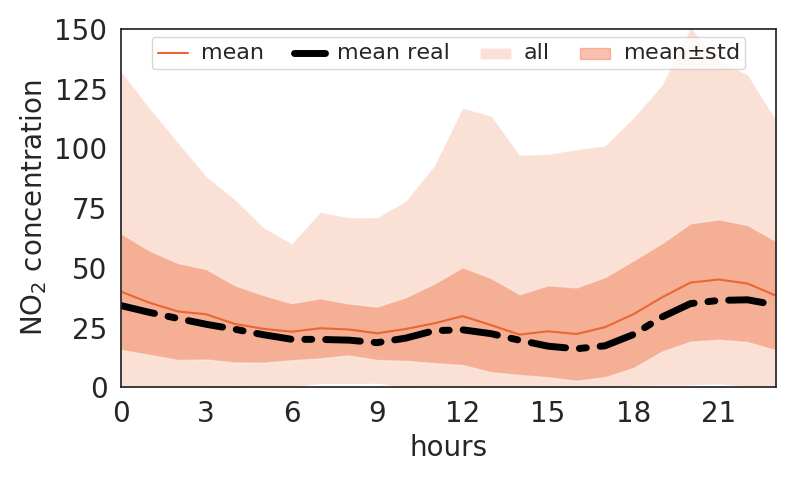}
    \footnotesize{a1) Class 0: winter-weekends.}
\end{minipage}
\begin{minipage}[l]{0.48\textwidth}
\centering
    \includegraphics[width=\textwidth]{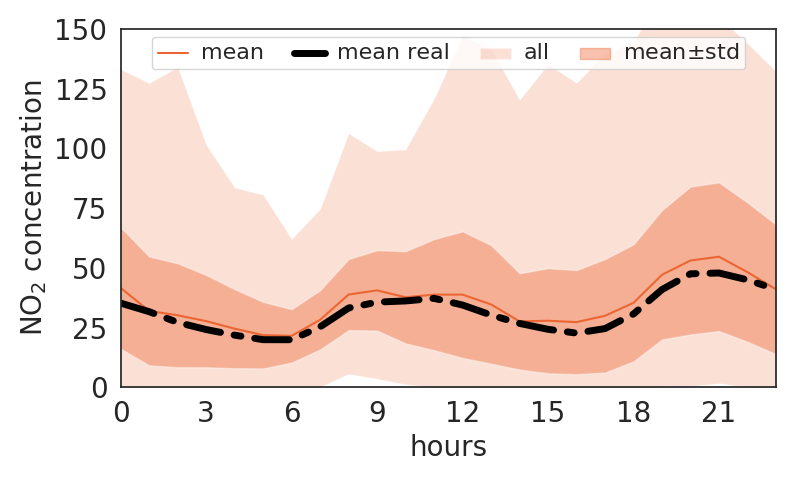}
    \footnotesize{a2)  Class 1: winter-working.} 
\end{minipage}
\vspace{0.2cm}

\begin{minipage}[l]{0.48\textwidth}
\centering
    \includegraphics[width=\textwidth]{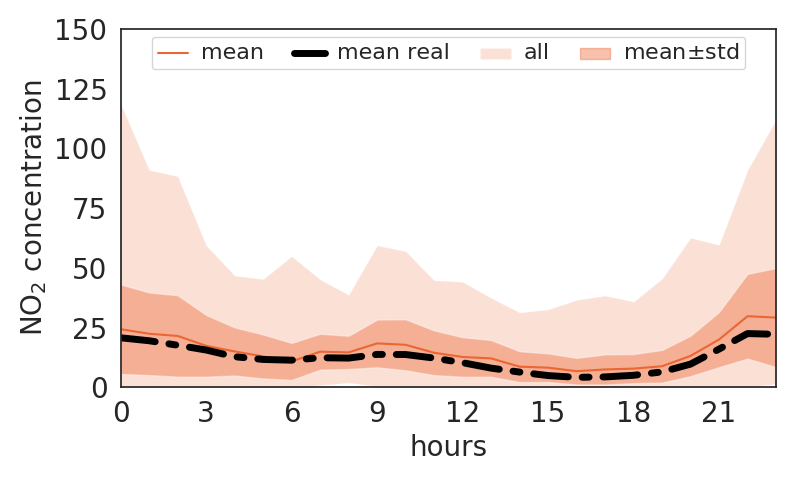}
    \footnotesize{b1) Class 2: spring-weekends.}
\end{minipage} 
\begin{minipage}[l]{0.48\textwidth}
\centering
    \includegraphics[width=\textwidth]{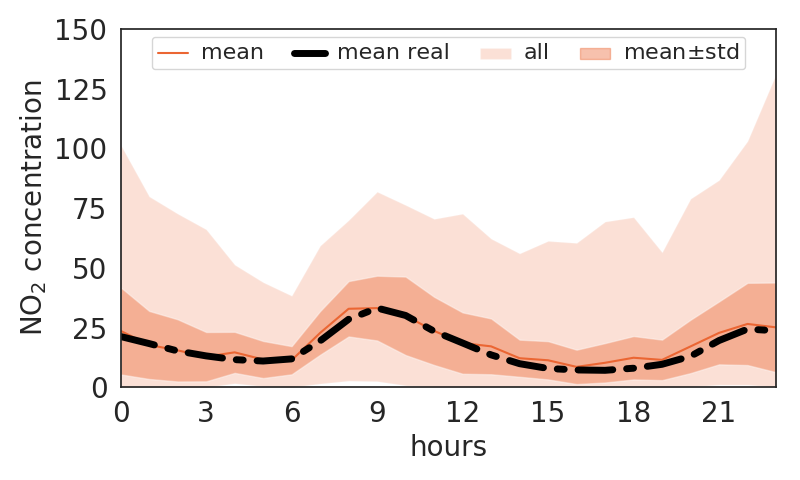}
    \footnotesize{b2) Class 3: spring-working.} 
\end{minipage} \\
\vspace{0.2cm}

\begin{minipage}[l]{0.48\textwidth}
\centering
    \includegraphics[width=\textwidth]{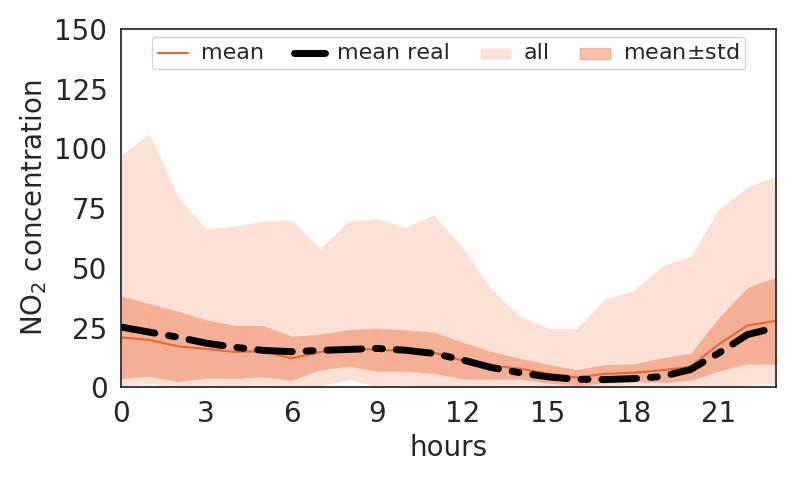}
    \footnotesize{c1) Class 4: summer-weekends.}
\end{minipage} 
\begin{minipage}[l]{0.48\textwidth}
\centering
    \includegraphics[width=\textwidth]{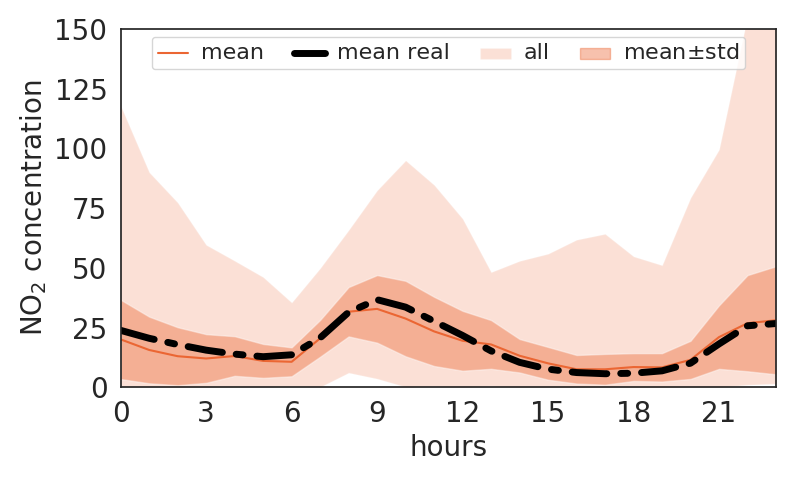}
    \footnotesize{c2) Class 5: summer-working.} 
\end{minipage} 

\begin{minipage}[l]{0.48\textwidth}
\centering
    \includegraphics[width=\textwidth]{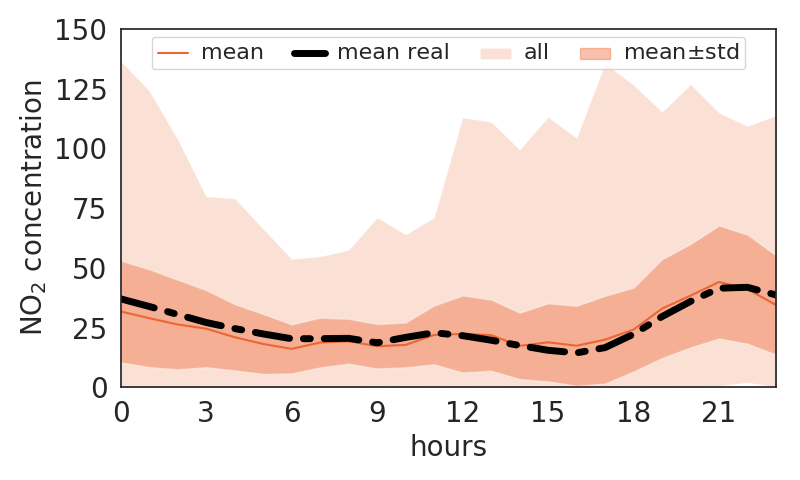}
    \footnotesize{d1) Class 6: autumn-weekends.}
\end{minipage} 
\begin{minipage}[l]{0.48\textwidth}
\centering
    \includegraphics[width=\textwidth]{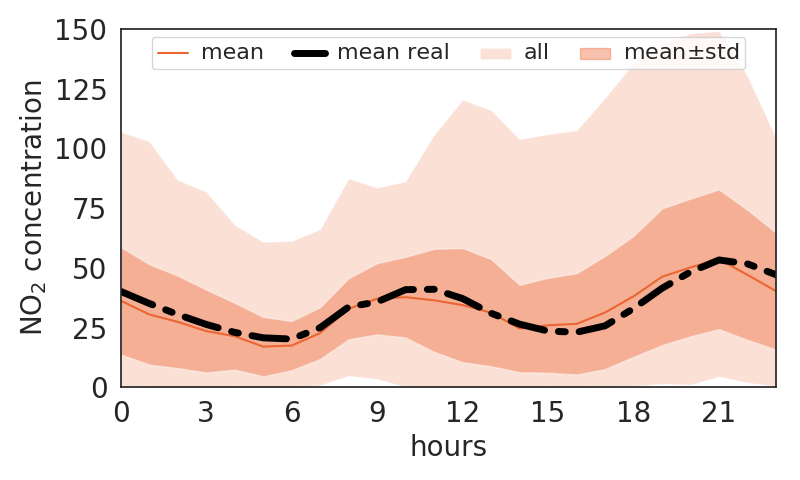}
    \footnotesize{d2) Class 7: autumn-working.} 
\end{minipage} 

\caption{Synthesized data samples created by \textit{fake-1}.}
\label{fig:fake_data_best1}
\end{figure}

\begin{figure}
\centering
\begin{minipage}[l]{0.48\textwidth}
\centering
    \includegraphics[width=\textwidth]{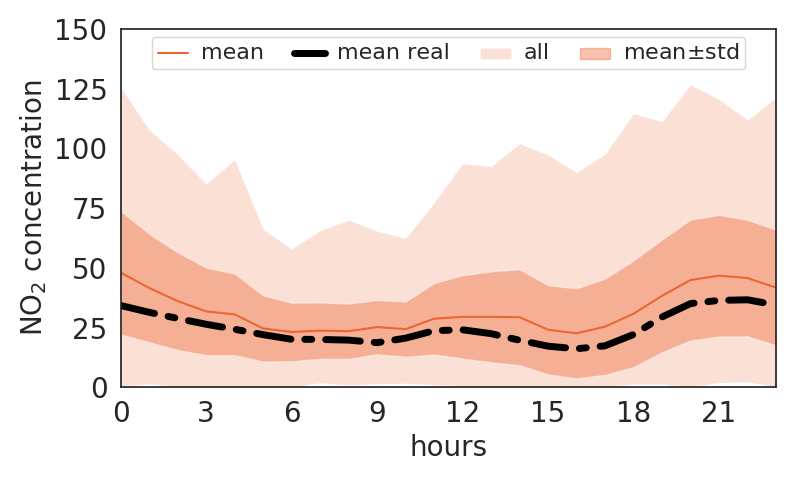}
    \footnotesize{a1) Class 0: winter-weekends.}
\end{minipage}
\begin{minipage}[l]{0.48\textwidth}
\centering
    \includegraphics[width=\textwidth]{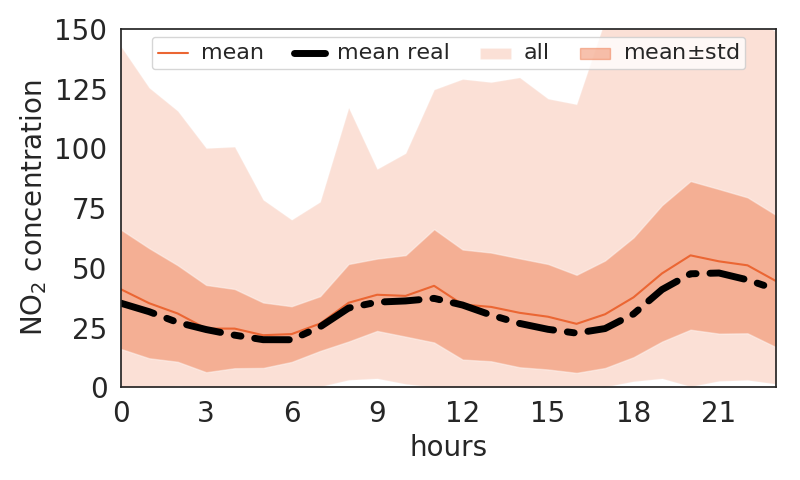}
    \footnotesize{a2)  Class 1: winter-working.} 
\end{minipage}
\vspace{0.2cm}

\begin{minipage}[l]{0.48\textwidth}
\centering
    \includegraphics[width=\textwidth]{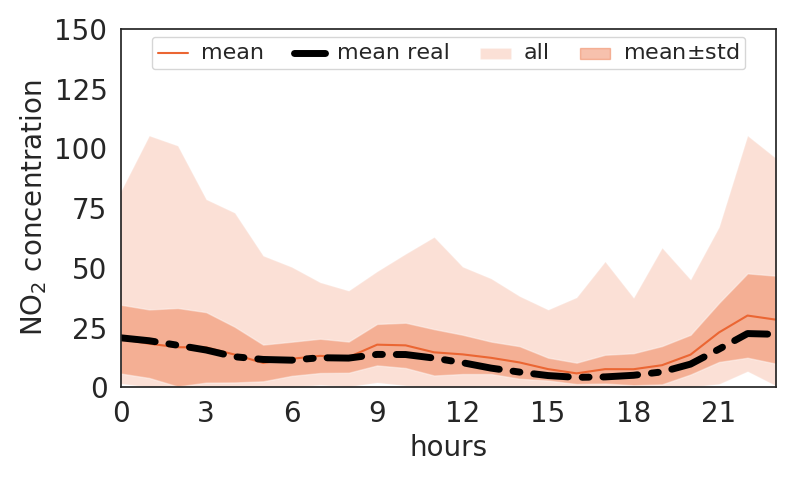}
    \footnotesize{b1) Class 2: spring-weekends.}
\end{minipage} 
\begin{minipage}[l]{0.48\textwidth}
\centering
    \includegraphics[width=\textwidth]{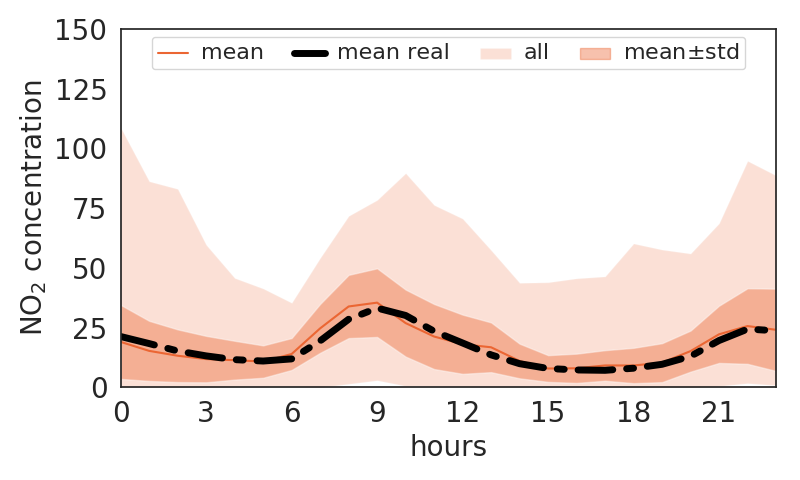}
    \footnotesize{b2) Class 3: spring-working.} 
\end{minipage} \\
\vspace{0.2cm}

\begin{minipage}[l]{0.48\textwidth}
\centering
    \includegraphics[width=\textwidth]{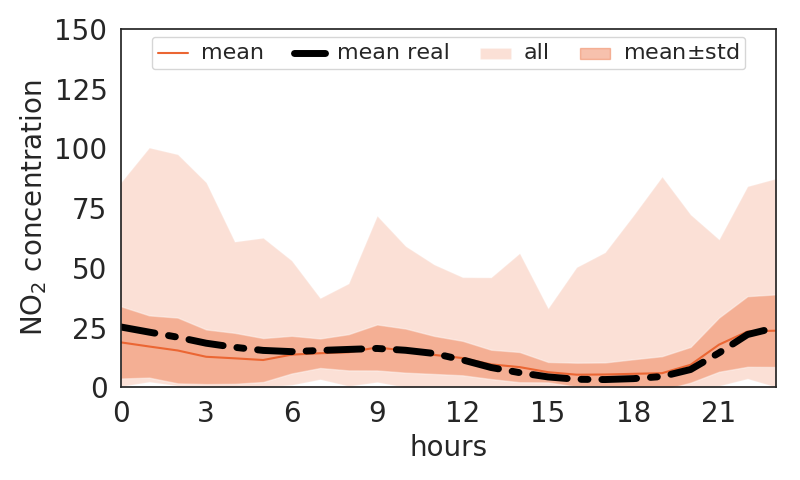}
    \footnotesize{c1) Class 4: summer-weekends.}
\end{minipage} 
\begin{minipage}[l]{0.48\textwidth}
\centering
    \includegraphics[width=\textwidth]{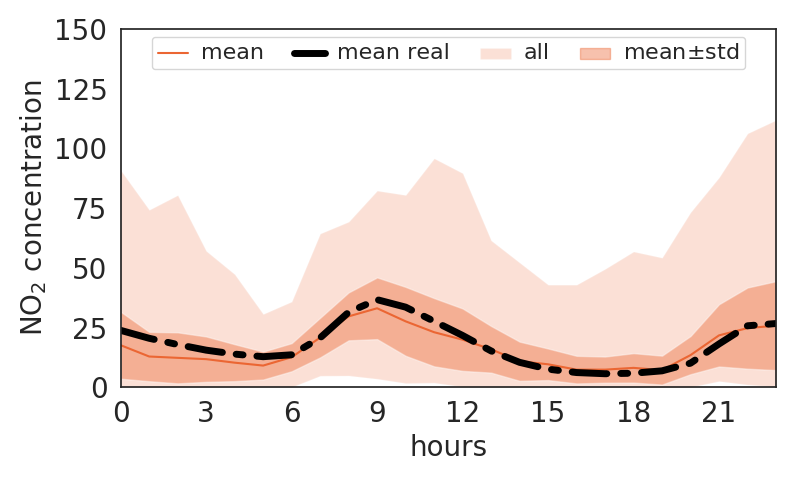}
    \footnotesize{c2) Class 5: summer-working.} 
\end{minipage} 

\begin{minipage}[l]{0.48\textwidth}
\centering
    \includegraphics[width=\textwidth]{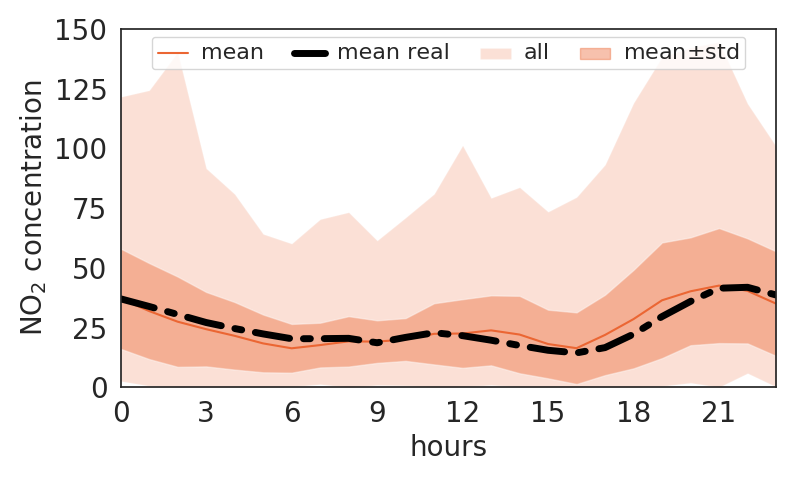}
    \footnotesize{d1) Class 6: autumn-weekends.}
\end{minipage} 
\begin{minipage}[l]{0.48\textwidth}
\centering
    \includegraphics[width=\textwidth]{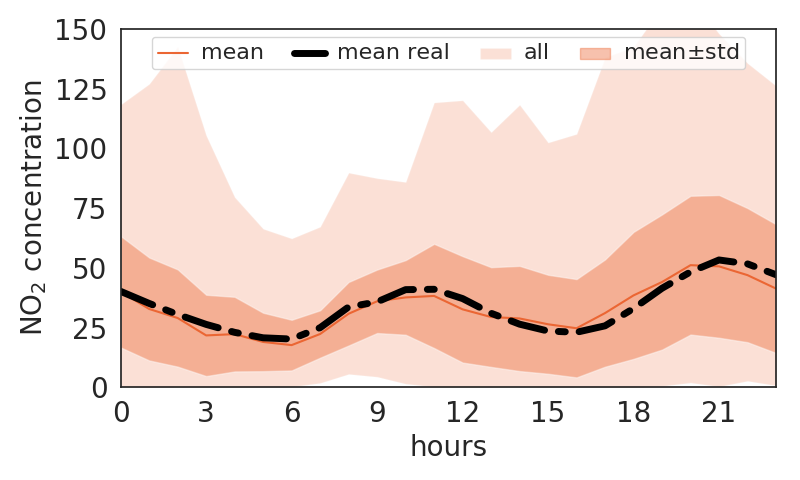}
    \footnotesize{d2) Class 7: autumn-working.} 
\end{minipage} 

\caption{Synthesized data samples created by \textit{fake-10}.}
\label{fig:fake_data_best3}
\end{figure}

The robustness shown by the GAN training when addressing the problem studied here allows the generation of realistic synthetic daily pollution time series. 
Here, we illustrate the 40,000 samples generated that belong to \textit{fake-1} and \textit{fake-10} datasets, the fake datasets with the best and the worst RMSE. 
Figs.~\ref{fig:fake_data_best1} and~\ref{fig:fake_data_best3} summarizes the samples by showing the mean values (the orange line), 
the values between the mean $\pm$ the standard deviation (the dark orange area), 
and all other values, i.e., between the minimum and maximum (the lighter orange area). In turn, the dotted black line presents the mean value of the real training data set (see Fig~\ref{fig:real_data}).

As can be seen in both figures, the mean values of the fake data and the mean values of the real data are very close. However, the distance between the orange line and the dark line increases for \textit{fake-10}. 

The shape of the dark orange areas are different depending on the class, but similar among the different datasets, i.e., 
the three generators studied produce samples with similar general trends. However, the shapes are less wide for the \textit{fake-1} dataset (it shows the lowest standard deviation in Table~\ref{tab:rmse-total}). This is mainly because it tends to generate less diverse samples than the other evaluated generators. 
Notice, that the fake datasets are ranked taking into account the RMSE against the mean curves of the real data set. 

Thus, according to the insights got from the datasets illustrated in Figs.~\ref{fig:fake_data_best1} and~\ref{fig:fake_data_best3} and the results in tables~~\ref{tab:rmse-total} and \ref{tab:rmse-class}, we can see the effectiveness of the methodology proposed to augment the pollution daily time series; and therefore the answer to \textbf{RQ}: Is it possible to apply generative modeling to produce new daily time series to improve our understanding of the phenomena related to the pollution in our cities? is yes.

\subsubsection{Computational time}

One of the main drawbacks of applying ANNs and deep learning approaches, as CGANs, is the computational effort (time and memory) required to train the models. 
As the size of the data samples and the ANN models used are not significant, the memory is not an issue et all. 
Regarding the computational times, the minimum, mean, and maximum times were 67.68, 69.64, and 72.04~minutes, respectively. 
That represents a non-very-high time-consuming investment, mainly because once the generators are trained, they create the new samples instantaneously.  

\newpage
\section{Conclusions and future work}
\label{sec:conclusions}

The interest in modeling, predicting, and forecasting ambient air pollution has been growing during the last years. Data-driven methods suffer from a lack of data to provide more accurate results. Thus, we propose the use of CGANs to train generative models able to create synthesized daily time series of a given pollutant from a given area of a city according to a given label. In this research, we have modeled NO$_2$ concentration at the downtown of Madrid as a use case.

The main results indicate that the proposed model is able to generate accurate NO$_2$ pollution daily time series while requiring a reduced computational time.
CGANs have shown robustness on the training because all the experiments converged to accurate generators. 
Thus, we are optimistic that this is the first step to develop more complex generative models able to produce synthetic pollution of a whole city taking into account information as the weather or the road traffic. 

The main lines for future work are related to 
extend the proposed model to generate the pollution of the whole city of Madrid by taking into account information from different sensors,
propose other classification that will allow including road traffic density and the weather (it will require the definition of more classes), 
and applying the generated data to feed data-driven models to prove that they are able to improve their accuracy after including fake samples. 


\bibliographystyle{splncs04}
\bibliography{refs}
\end{document}